\documentclass[runningheads]{llncs}

\usepackage[T1]{fontenc}

\usepackage{booktabs}
\usepackage{graphicx}
\usepackage{xcolor}
\usepackage{colortbl}
\usepackage{diagbox}
\usepackage[normalem]{ulem}

\definecolor{ClassifierBlue}{HTML}{0057B8}
\definecolor{AgentOrange}{HTML}{B54708}

\usepackage{newtxtext}       
\usepackage[varvw]{newtxmath}

\usepackage{graphicx}

\usepackage{comment}
\usepackage{todonotes}
\usepackage{tikz}
\usetikzlibrary{shapes.geometric, arrows.meta, positioning, fit, backgrounds, calc}
\usepackage{graphicx} 

\begin{document}

\title{JOVE-ICE: Intent-Guided Cloud–Edge–Device Collaboration for AI Glasses}

\author{Yuyang Jin\inst{1} \and
Bingfeng Pi\inst{1} \thanks{Corresponding author: pibf@think-land.com} \and
Jiamei Sun \inst{1} \and
Wei Jin \inst{2} \and
Dongyang Wang\inst{3}\and
Yunpeng Li\inst{4} \and
Yan Bai \inst{2} \and
Lei Zuo \inst{1} 
}

\authorrunning{Y. Jin et al.}

\institute{
TianjuDihe (Suzhou) Technology Co., Ltd.\\
\email{jinyuyang@think-land.com, pibf@think-land.com, sunjm@think-land.com, lei.zuo@think-land.com}\\
\and
Suzhou Jupiter Smart Technology Co., Ltd. \\
\email{jinwei@szjove.cn, baiyan@szjove.cn} \\
\and
Monash University Joint Graduate School (Suzhou), Southeast University\\
\email{220255090@seu.edu.cn}
\and
Xi'an Jiaotong-Liverpool University \\
\email{Yunpeng.Li21@student.xjtlu.edu.cn}
}

\maketitle              

\begin{abstract}
Recent advances in large language models (LLMs) and LLM-based agents have expanded the capabilities of AI glasses, enabling applications such as egocentric visual question answering, long-term memory retrieval, and proactive assistance. AI glasses process multimodal inputs, including speech, images, and video, to support tasks ranging from device control and text-based interactions to visual understanding. This diversity of inputs and tasks creates substantial computational and communication demands, making efficient deployment at scale a key challenge. Although video-centric capabilities have received considerable attention, simpler text-only and device-executable tasks may account for a substantial share of everyday interactions and remain less explored. Supporting this diverse workload within a unified and efficient assistant is an open challenge.
We propose JOVE-\textbf{ICE}, an \textbf{I}ntent-guided \textbf{C}loud–\textbf{E}dge–device collaborative framework that routes requests among AI glasses, a paired smartphone, and cloud servers according to user intent. The framework combines hierarchical intent classification with dynamic subagent orchestration to select suitable execution paths. This orchestration mechanism makes JOVE-ICE flexible and extensible, allowing it to support new applications through the integration of additional specialized subagents.
We collect real-world AI-glasses requests and find that simple device-control tasks account for more than \textbf{30\%} of interactions. JOVE-ICE handles these requests on-device, avoiding unnecessary cloud use and associated costs.
We further evaluate JOVE-ICE on three public benchmarks. Compared with the commonly used device–cloud framework, JOVE-ICE reduces \textit{time to first token} (TTFT) by approximately \textbf{13\%} and token usage by more than\textbf{ 11\%} while maintaining task quality. These results highlight the potential of JOVE-ICE to improve the efficiency of AI glasses and support their practical deployment at scale.

\end{abstract}

\section{Introduction}
AI glasses are emerging as part of a new wearable computing ecosystem~\cite{yuan2026aismartglasseswearable}, however, their computing, battery, and network resources are substantially more constrained than those of smartphones. At the same time, users expect natural interactions with conversational latency, making delays caused by cloud-dependent processing a significant barrier to practical use. Designed for continuous wear, AI glasses incur added latency and energy costs with each unnecessary uplink.

Existing AI-glasses frameworks have largely focused on expanding functionality, with less attention to computational overhead, bandwidth consumption, and their effects on end-to-end latency. In parallel, research has primarily emphasized video-centric capabilities~\cite{sitong2026vinci2providingproactiveassistance,vinci,liu2026visionclawalwaysonaiagents,yang2026egocentriccopilotwebnativesmartglasses}, which are central to AI glasses. However, AI glasses can also provide a hands-free, multimodal interface—combining audio, text, video, and potentially other sensor data—for personal-assistant tasks such as device control, app interactions, casual conversation, web search, and knowledge base or database retrieval ~\cite{baek2025knowledgebaseconstructionknowledgeaugmented,brahim2026constructionnaturallanguagequerying}. Integrating these diverse functions into a single, efficient assistant remains an open challenge. This paper focuses on two related challenges: \textbf{(C1)} improving efficiency under device and communication constraints, and \textbf{(C2)} supporting diverse personal-assistant functions within a unified, extensible framework.

To address these challenges, we propose JOVE-ICE, an intent-guided cloud–edge–\allowbreak device collaborative framework for AI glasses. JOVE-ICE is designed to achieve five key objectives: \textbf{(G1)} immediate feedback for frequent commands, \textbf{(G2)} minimal bandwidth and energy consumption, \textbf{(G3)} cloud-comparable task quality, \textbf{(G4)} extensibility via pluggable subagents, and \textbf{(G5)} portability across heterogeneous hardware. 

To realize these goals, JOVE-ICE employs a three-tier cascade that uses intent as a routing signal to match each request with the least resource-intensive execution tier capable of handling it. A lightweight acoustic front-end on the glasses ensures low-latency local processing (addressing \textbf{G1} and \textbf{G2}), while deterministic matching on the paired smartphone and neural classification in the cloud handle increasingly complex tasks. This hierarchy keeps routine controls local and escalates implicit or knowledge-intensive requests to more capable models, ensuring high accuracy \textbf{(G3 of high quality)}. Furthermore, predicted intents route requests to specialized, pluggable subagents (e.g., small talk, visual, or tool-augmented), supported by a session manager that maintains context and detects boundaries to limit growth \textbf{(G4 of extensibility)}. Finally, the framework adapts to varying hardware capabilities through a device-specific capability profile, ensuring seamless deployment across diverse form factors such as audio-only (W1), display-equipped (L1), and camera-enabled (S1) glasses \textbf{(G5 of portability)}.

We quantitatively evaluate JOVE-ICE on three public benchmarks covering visual question answering and text-only interactions. We further assess its effectiveness using two private datasets derived from real-world AI-glasses requests. The results show that JOVE-ICE reduces interaction latency and token usage while maintaining task quality.
Our main contributions are: 
\begin{itemize} 
\item[$\bullet$] We present JOVE-ICE, a cloud--edge--device collaborative framework that adapts to AI glasses with different hardware capabilities and can be extended to support new applications.
\item[$\bullet$] We introduce a hierarchical intent-classification cascade and dynamic subagent orchestration mechanism that route tasks based on user intent, allocating computation across the device, edge, and cloud to improve efficiency and reduce cloud usage.
\item[$\bullet$] We conduct extensive experiments on public benchmarks and real-world AI-glasses request datasets. Experiments demonstrate that intent-guided resource allocation reduces latency by approximately 13\% without compromising performance, validating its potential for large-scale, practical deployment.
\end{itemize}

\section{Related Works}
Existing AI-glasses systems have evolved from reactive assistants toward continuous perception and proactive assistance. Recent systems such as VisionClaw and Vinci2 combine egocentric context with agent capabilities, including long-term video processing and proactive intervention~\cite{sitong2026vinci2providingproactiveassistance,liu2026visionclawalwaysonaiagents}. However, efficiently coordinating computation across glasses, smartphones, and cloud services remains an open challenge~\cite{yuan2026aismartglasseswearable}. This paper addresses the complementary issues of computational efficiency and economic viability in AI-glasses systems. 

For intent recognition, mainstream research has focused on clarifying ambiguous requests via interactive dialogue or direct LLM inference~\cite{qian2024tellmoreimplicituser}, while other work leverages intent mainly to guide data transmission between glasses and the cloud~\cite{jiang2026intentionawaresemanticagentcommunications}. In contrast, we use recognized intent to dynamically route tasks to the most appropriate execution path across the device–edge–cloud continuum.

Most existing AI glasses adopt a device–cloud framework~\cite{jiang2026intentionawaresemanticagentcommunications,yang2026egocentriccopilotwebnativesmartglasses}. VisionClaw~\cite{liu2026visionclawalwaysonaiagents} extends this to a device–edge–cloud setting by incorporating the smartphone, yet the phone primarily serves data streaming rather than computation. Differently, we jointly leverage the computational resources of the glasses, smartphone, and cloud to balance tasks across the three tiers.

\section{Method}

\begin{figure}[t]
\includegraphics[width=0.95\textwidth]{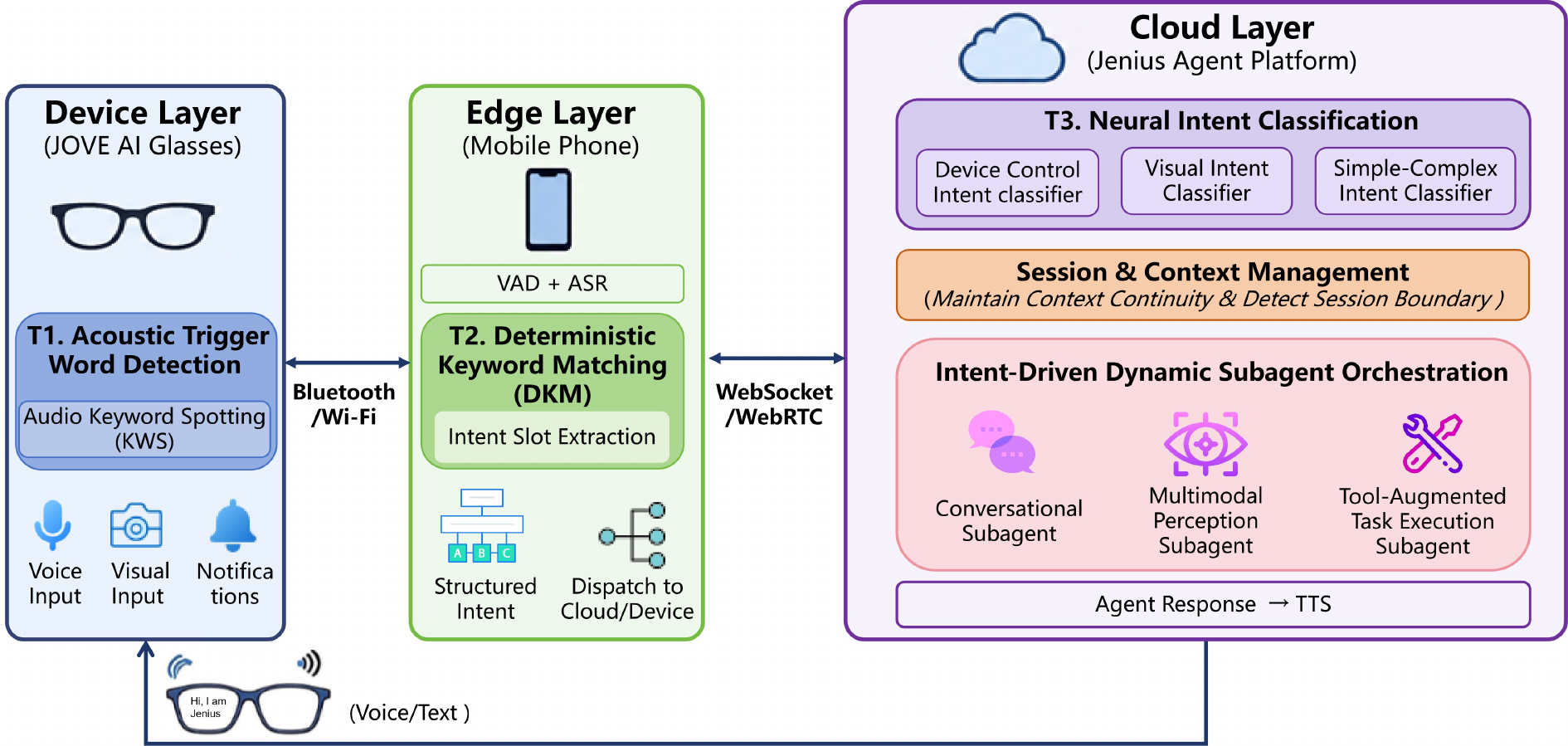}
\caption{Overview of JOVE-ICE, an intent-guided cloud-edge-device collaboration framework for AI glasses.}
\label{fig:framework}
\end{figure}

Let $q$ denote the user query; the routing policy $\pi$ determines the execution tier $L \in \{D, E, C\}$ (corresponding to the glasses device, the mobile edge, and the cloud) and the execution entity $A$ (e.g., control command executed on device side, task executed with a specific agent in the cloud ) based on intent classification. Our optimization objective is to minimize the end-to-end interaction latency and the cost of model invocation, subject to a quality constraint. This can be formally expressed as:
\begin{equation}
    \min\mathbb{({E}}[T_{\text{e2e}}(q,\pi)]+\lambda \mathbb{E}[C_{\text{token}}(q,\pi)]), \text{subject to}: {Q(q,\pi) \ge Q_\text{base}(q)-\epsilon}
    \label{equ:expected_cost}
\end{equation}
where $T_{\text{e2e}}(q,\pi)$ represents the total end-to-end latency, encompassing local detection on device and phone, network transmission, intent classification, model generation, and response delivery (text/voice); $C_{\text{token}}(q,\pi)$ denotes the total number of tokens consumed in a single interaction; $Q(q,\pi)$ is the task quality metric, while $Q_\text{base}(q)$ is the quality achieved by a baseline method without intent-based routing; $\epsilon$ is the upper bound of acceptable quality degradation; $\lambda$ is a weighting factor that balances the trade-off between latency and token cost. This formulation highlights that the value of our routing mechanism lies not merely in offloading tasks to the edge, but in avoiding unnecessary cross-tier transmissions and excessive model inference while strictly adhering to quality constraints.

Therefore, we propose JOVE-ICE, an intent-guided cloud–edge–device collaborative framework designed to balance latency, efficiency, and task quality for AI glasses. As illustrated in Fig.~\ref{fig:framework}, JOVE-ICE employs a three-tier cascade that progressively enriches processing capabilities while minimizing resource consumption. The core principle is to use intent as a routing signal, matching each request to the least resource-intensive execution tier capable of handling it, and escalating only when necessary.
The data flow proceeds as follows:

\textbf{Device Tier (On-Glasses)}: Incoming audio is first processed by a lightweight, always-on Acoustic Trigger Word Detection module. This ensures immediate feedback for frequent, unambiguous commands with minimal energy consumption.

\textbf{Edge Tier (Paired Smartphone)}: Audio segments that miss the on-device trigger are forwarded to the paired smartphone via Bluetooth or Wi-Fi. Here, Deterministic Keyword Matching operates on the ASR transcripts to handle structured device-control commands locally, avoiding unnecessary cloud transmission.

\textbf{Cloud Tier}: Unmatched or complex tasks are offloaded to the cloud via WebSocket (prototype) or WebRTC (production). In this tier, Neural Intent Classification predicts the user’s intent which then routes the request to specialized, session-aware multimodal subagents. These subagents handle diverse tasks, including visual QA, text-based QA, and tool-augmented interactions. Notably, image frames are transmitted on-demand only for requests predicted as \textbf{NEED\_IMAGE} and only from camera-equipped devices, thereby preserving bandwidth. A dedicated Session Manager maintains context across subagents and detects session boundaries to limit context growth.

The remainder of this section details the three core components of JOVE-ICE.

\subsection{Hierarchical Intent Classification}
\label{sec:Intent_classification}

\begin{table}[t]
\centering
\small
\setlength{\tabcolsep}{4pt}
\caption{Three-tier hierarchical intent classification.}
\label{tab:cascade}
\begin{tabular}{@{}p{2.2cm}p{2.6cm}p{2.6cm}p{2.6cm}@{}}
\toprule
& \textbf{Tier-1 (Glasses)} & \textbf{Tier-2 (Phone)} & \textbf{Tier-3 (Cloud)} \\
\midrule
Method & CNN KWS & Template matching & MacBERT/XLM-R \\
Input  & Raw PCM audio & ASR transcript & ASR transcript \\
Coverage & 6 Cmds. & 11 Cmds. + other & 11 Cmds. + other \\
Escalates when & $P_{\text{trigger}} < \theta_{\text{kws}}$ & No template match & Always (last tier) \\
\bottomrule
\end{tabular}
\end{table}

To achieve low interaction latency while preserving system power and transmission bandwidth, we design a three-tier, hierarchical intent classification framework that dynamically offloads computation across the AI glasses, the paired smartphone gateway, and the cloud server, effectively balancing efficiency and capability. \textbf{T1} handles frequent AI glasses commands locally for minimal latency; \textbf{T2} matches Automatic Speech Recognition (ASR) transcripts against extensible templates with negligible cost; and \textbf{T3} resolves complex, implicit intents via cloud semantics. This design minimizes network usage and preserves privacy for most interactions, while ensuring graceful degradation in offline scenarios. The three-tier cascading mechanism is detailed in Table~\ref{tab:cascade}.

\textbf{T1. Acoustic Trigger Word Detection (On-Device)}
The frontline trigger word detection module is embedded directly on the smart glasses to perform lightweight, compact Convolutional Neural Network (CNN) keyword spotting (KWS) over log-Mel features~\cite{sainath2015cnn}. The distinctiveness of JOVE-ICE lies not in its front-end architecture, but in its operational contract within the cascade. This contract comprises three key constraints: (1) a closed vocabulary of six fixed triggers \footnote{play music, pause music, previous track, next track, volume up, volume down}; (2) a conservatively tuned activation threshold $\theta_{\text{kws}}$; and (3) a strict execution policy where local hits are processed entirely on-device without network transmission, while misses trigger audio streaming to the paired phone. By maintaining a small, closed vocabulary, the keyword spotter achieves the computational efficiency required to reside on an always-on MCU/DSP, thereby minimizing power consumption.

\textbf{T2. Deterministic Keyword Matching (Mobile Gateway)}
When an audio segment bypasses T1, it is streamed to the paired smartphone via Wi-Fi or Bluetooth, depending on the communication capability of the glasses. On the mobile edge gateway, the audio stream undergoes Voice Activity Detection (VAD) followed by ASR, producing a text transcript $T$. The gateway executes deterministic string matching against a predefined dictionary of 11 device-control commands $\mathcal{K}$ \footnote{open app, call contact, send SMS, look up phone number, take photo, record video, view calendar, set alarm, volume up, volume down, go to home page}:
\begin{equation}
    f_{\text{match}}(T, \mathcal{K}) = 
    \begin{cases} 
      c_i, & \exists k_i \in \mathcal{K} \text{ s.t. } k_i \subseteq T \\
      \emptyset, & \text{otherwise}
    \end{cases}
\end{equation}
If a match $c_i$ is established, the corresponding executable command is immediately transmitted back to the AI glasses over the Bluetooth channel, completely bypassing cloud-side inference overhead. Because $\mathcal{K}$ is plain text, coverage can be extended without retraining an acoustic model.

\textbf{T3. Neural Intent Classification (Cloud-based)}
Transcripts $T$ that remain unmatched at the mobile gateway are forwarded
to the cloud server to be assessed by three independently trained classifiers.

\textit{Device-Control Intent Classifier} employs a shared MacBERT-base encoder~\cite{cui2020macbert} to jointly predict device-control intents and extract corresponding parameters for the incoming transcript $T$. It comprises two prediction heads: an intent head that classifies inputs into one of 12 categories (11 specific device-control commands plus \texttt{OTHER}) via $\hat{y}_d=\arg\max_k p_d(k)$, and a slot head that extracts structured parameters, such as application names, contacts, and temporal expressions.

To ensure execution consistency, decoded slot spans are validated against the predicted intent before populating command templates. For instance, given the utterance “Wake me up at 7 AM”, the model predicts the intent \texttt{set alarm} and extracts the time slot \texttt{07:00am}. This structured output directly maps to the predefined command template, bypassing the need for complex subagent reasoning.

\textit{Visual Intent Classifier }determines whether the Transcript $T$ requires access to user's visual surroundings. XLM-R-base~\cite{conneau2020xlm-r}
encodes $T$, its first-token representation predicts \texttt{NEED\_IMAGE} / \texttt{NO\_NEED\_IMAGE}:
\begin{equation}
    \hat{y}_v =
    \begin{cases}
        \texttt{NEED\_IMAGE},
        & p_v(\texttt{NEED\_IMAGE}\mid T)\geq\gamma,\\
        \texttt{NO\_NEED\_IMAGE},
        & \text{otherwise},
    \end{cases}
\end{equation}
where $\gamma$ is a development-selected threshold. This prediction
gates frame capture on camera-equipped devices.

\textit{Simple-Complex Intent Classifier} distinguishes conversational requests from tasks
requiring tool use or external operations. MacBERT-large~\cite{cui2020macbert} encodes
the transcript, using its pooled sequence representation. The classification head predicts probabilities
over \texttt{SIMPLE} /
\texttt{COMPLEX}.
The highest-probability class,
$\hat{y}_c=\arg\max_k p_c(k)$, informs routing between conversational
response generation and tool-assisted task processing.

\subsection{Session Management}
\label{sec:session_mgt}
\begin{figure}[t]
\centering
    \begin{minipage}[t]{0.48\textwidth}
        \centering
        \includegraphics[width=\textwidth]{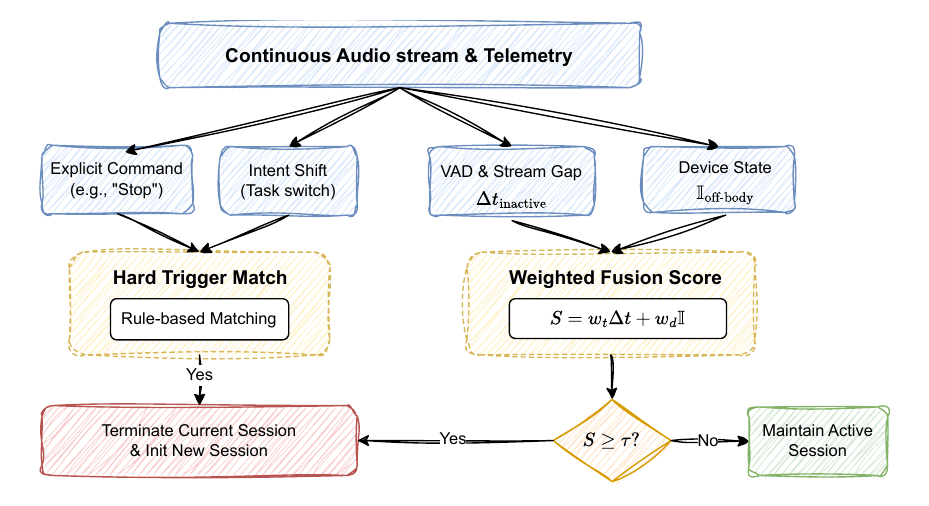}
        \caption{Multi-Signal Session Boundary Detection Pipeline.}
        \label{fig:session_boundary}
    \end{minipage}
    \hfill
    \begin{minipage}[t]{0.48\textwidth}
        \centering
        \includegraphics[width=\textwidth]{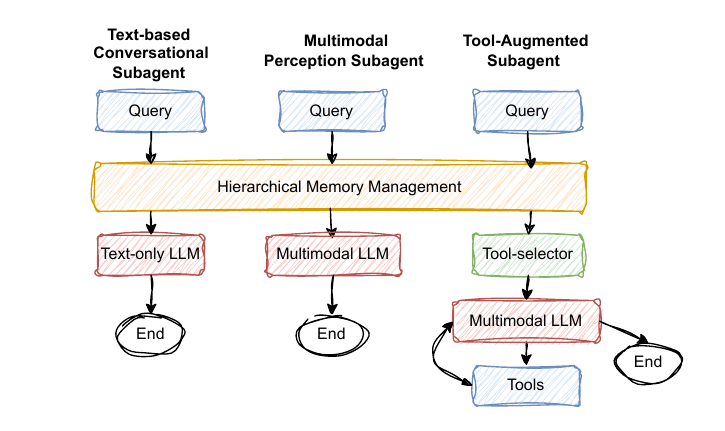}
        \caption{Implementation of task-specific subagents.}
        \label{fig:subagent}        
    \end{minipage}
\end{figure}

Concatenating continuous multimodal interactions into one unbounded context
window introduces visual/textual redundancy, growing memory overhead, and
unpredictable inference latency \cite{kwon2023efficient}; aggressive truncation, on the other hand, breaks semantic continuity. We combine \textbf{Hard Triggers} (deterministic termination) with \textbf{Soft Triggers} (weighted context fusion), as illustrated in Fig. \ref{fig:session_boundary}.

\textbf{Hard Triggers} ends a session immediately on two conditions. 1) \textbf{Explicit Termination Commands:} Semantic matching against a predefined set of explicit user commands (e.g., ``End chat,'' ``Stop''). 2) 
\textbf{Abrupt Intent Shift:} A sharp disjunction in intent classification, such as a transition from open-domain QA to low-level hardware control.

\textbf{Soft triggers} model session boundaries for ambiguous interactions using a weighted scoring function $S$ that combines temporal inactivity with hardware sensor telemetry:
\begin{equation}
\label{equ:soft_trigger}
    S = w_t \cdot \Delta t_{\text{inactive}} + w_d \cdot \mathbb{I}_{\text{off-body}}
\end{equation}
where $\Delta t_{\text{inactive}}$ represents the cumulative duration of voice inactivity (evaluated via VAD and stream inter-arrival gaps), $\mathbb{I}_{\text{off-body}} \in \{0, 1\}$ indicates whether the glasses have been removed, and $w_t, w_d$ are learned or calibrated weights. Device removal $\mathbb{I}_{\text{off-body}} = 1$ substantially inflates $S$, causing the score to cross threshold $\tau$ within a markedly narrower time window. When $S \ge \tau$, the current session is safely archived, and a new context is initialized for subsequent interactions.

\subsection{Intent-Driven Dynamic Subagent Orchestration}
\label{sec:agent_orchestration}
Built upon the Jenius agent platform~\cite{xia2026jeniusagent}, which enables developers to define, register, and orchestrate task-specific subagents, we implement an intent-guided subagent orchestration architecture for AI glasses. Leveraging the predictions from neural intent classification module, we develop three representative subagents to address diverse interaction requirements, as shown in Fig.\ref{fig:subagent}. Each subagent employs the hierarchical memory management approach introduced in~\cite{xia2026jeniusagent}.

\textbf{Text-based Conversational Subagent} is activated when the device-control intent classifier outputs $\hat{y}_d = \texttt{OTHER}$, the visual-intent classifier predicts $\hat{y}_v = \texttt{NO\_NEED\_IMAGE}$, and the complexity classifier yields $\hat{y}_c = \texttt{SIMPLE}$. It handles lightweight tasks, such as small talk and general dialogue, by combining the query with retrieved context and a lightweight text-only LLM.

\textbf{Multimodal Perception Subagent} is activated when $\hat{y}_d = \texttt{OTHER}$, $\hat{y}_v = \texttt{NEED\_IMAGE}$, and $\hat{y}_c = \texttt{SIMPLE}$. It
performs visual question-answering tasks through direct visual understanding with a multimodal LLM, without relying on external tools.

\textbf{Tool-Augmented Subagent} is activated when $\hat{y}_d = \texttt{OTHER}$ and $\hat{y}_c = \texttt{COMPLEX}$ (regardless of $\hat{y}_v$). It supports external retrieval and service interactions via a multimodal LLM-driven iterative tool invocation loop, integrating skills, APIs, MCPs, and fixed Robotic Process Automation (RPA) workflows to ensure task completion through dynamic context-aware reasoning.

Final responses are converted to speech by the module of Text-to-Speech (TTS), or rendered on the lens for display-equipped devices, and returned to the glasses.

\section{Experiments}
\subsection{Datasets}

We evaluate our framework on three public wearable-device datasets. \textbf{SuperGlasses} \cite{jiang2026superglasses} contains 2,422 image–question–answer samples spanning diverse domains and query types. \textbf{WearVQA} \cite{chang2026wearvqa} provides 2,500 visual question-answering samples captured in real-world egocentric settings. \textbf{WearVox} \cite{lin2026wearvox} covers five audio-based tasks, including search-grounded and closed-book question answering, side-talk rejection, tool calling, and speech translation. For the text-only evaluation, we employ a filtered subset of 4,260 instances from WearVox that possess non-empty reference transcripts. Collectively, these benchmarks enable a comprehensive assessment of visual and text understanding, reasoning capabilities, and task-specific response generation.

To evaluate the three-tier hierarchical intent-classification and session-management modules in real-world settings, we collected interaction data from users wearing three JOVE AI-glasses form factors: W1, L1, and S1. After removing invalid queries and duplicates, we compiled two private datasets. The \textbf{CascadeEval} comprises 428 queries derived from everyday usage scenarios. It evaluates the cascade’s potential to reduce resource consumption and latency by measuring the proportion of requests that are resolved locally on either the glasses or the paired smartphone. The \textbf{Session Management Dataset} comprises 199 manually annotated query turns from multi-turn interactions. Each turn is labeled as \textit{Split} (starting a new session; 34 turns, 17.1\%) or \textit{Continue} (continuing an existing session; 165 turns, 82.9\%).

\subsection{Cascade Tier Allocation Statistics}
\label{sec:deployment_stats}

\begin{figure}[t]
\centering
\begin{minipage}[c]{0.35\textwidth}
\centering
\includegraphics[width=\linewidth]{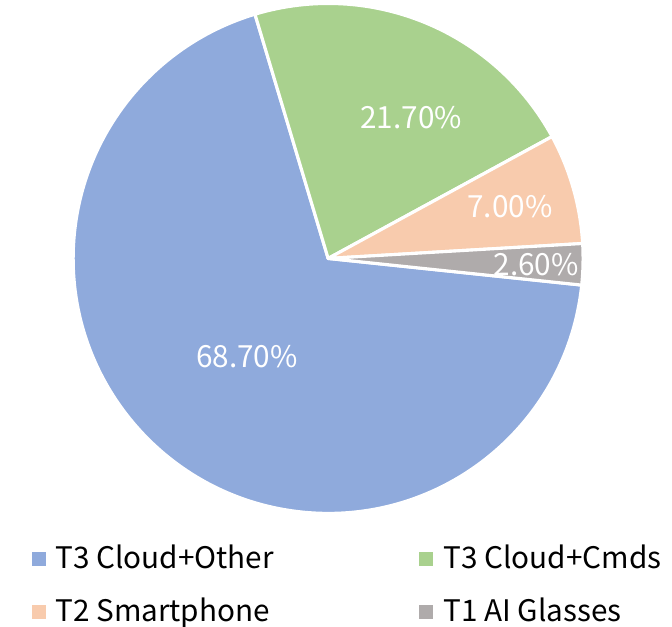}
\end{minipage}%
\hfill
\begin{minipage}[c]{0.62\textwidth}
\centering
\includegraphics[width=\linewidth]{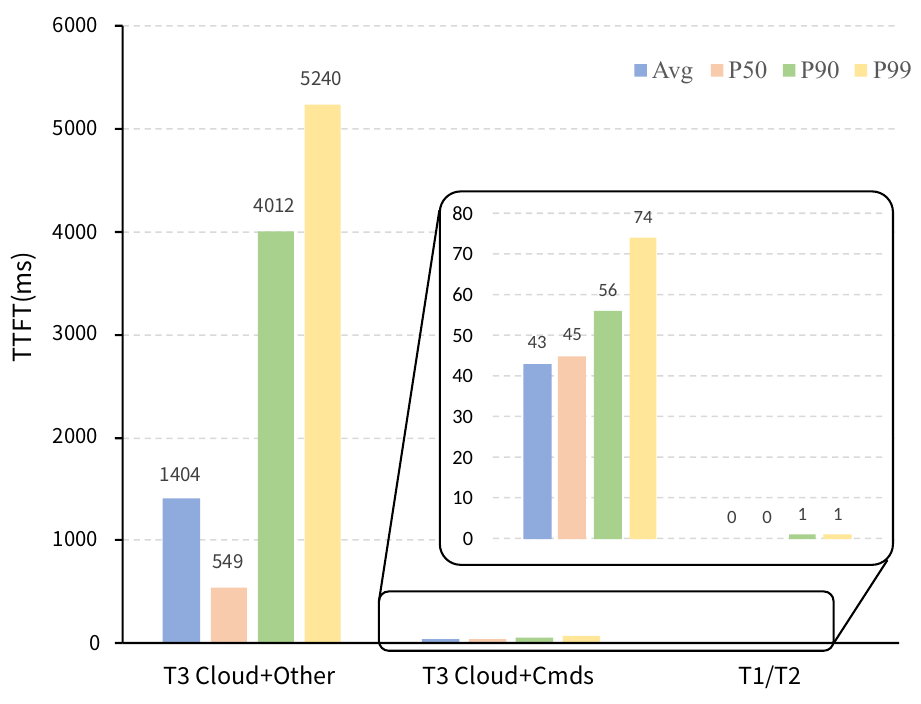}
\end{minipage}
\caption{Routing distribution (left) and TTFT percentiles by tier (right) on the CascadeEval dataset. TTFT is measured in \textbf{milliseconds} from query submission on the AI glasses to receipt of the first response token.}
\label{fig:routing_stats}
\end{figure}

Before examining the performance of JOVE-ICE, we first analyze the distribution of task types in real-world scenarios. We replay the 428 queries from the CascadeEval dataset on the JOVE AI glasses L1 and S1 to evaluate the framework’s end-to-end workflow. For each query, we record the processing tier and the \textbf{Time to First Token (TTFT)}, defined as the elapsed time until the first non-empty output segment is generated. 

Fig.~\ref{fig:routing_stats} summarizes the routing distribution and TTFT by tier; Tier T3 is further partitioned into device-control commands (Cmds.) and \texttt{OTHER} queries according to the device-control intent classifier. Our results reveal three key findings. First, device-control queries account for more than 30\% of requests across T1, T2, and T3 Cloud + Cmds., reflecting the prevalence of everyday actions such as opening an application, calling a contact, or sending a message. Second, 9.6\% of requests detected and directly handled by Tiers T1 and T2 exhibit near-zero latency. Third, at Tier T3, the cloud-based device-control classifier identifies 21.7\% of requests for execution on T2, achieving a median latency of 43 ms and a P99 latency of 74 ms. Together, these findings show that JOVE-ICE avoids unnecessary cloud-agent processing by routing device-control queries to the device side, enabling faster interactions and more efficient resource allocation. We will further evaluate the T3 Cloud+Other latency in the next section with public benchmarks.

\subsection{Visual Question Answering and Text-Only Interaction}
\label{sec:VQA_TextOnly}

In this section, we evaluate our framework’s task-execution capabilities on VQA and text-only interaction tasks using the \textbf{SuperGlasses}, \textbf{WearVQA}, and \textbf{WearVox} datasets. To match the benchmarks’ evaluation protocols, we run these experiments through the cloud layer, including the neural intent classification, session management, and dynamic subagent orchestration. The Text-based Conversational Subagent uses Qwen3.5-9B as its backbone language model, while the remaining subagents use Qwen3.8-27B. The three intent classifiers are jointly deployed across two RTX 4090 GPUs. We deploy Qwen3.8-27B on two 40-GB A100 GPUs and Qwen3.5-9B on one GPU.

We compare JOVE-ICE against the widely adopted device–cloud framework used in recent work~\cite{jiang2026intentionawaresemanticagentcommunications,yang2026egocentriccopilotwebnativesmartglasses}, which we denote as \textbf{Fixed Route (FR)}. In the FR baseline, all requests are routed exclusively to the Tool-Augmented Subagent. We evaluate the two approaches with respect to  response quality, latency, and token efficiency. Following each benchmark’s original evaluation protocol, we report \textbf{accuracy} on SuperGlasses and WearVQA. For WearVox, we report \textbf{mean quality scores} for dialogue translation and \textbf{accuracy} for the other four tasks.
To evaluate latency, we submit requests sequentially and measure two metrics: \textbf{TTFT}, as defined in Sec.~\ref{sec:deployment_stats}, and \textbf{end-to-end latency (E2E)}, defined as the time from request submission to task completion. For token usage, we report the mean number of input and output tokens per sample. For Neural Intent Classification, we estimate per-request latency as the longest GPU runtime among the three classifiers and average these estimates within each dataset.

Table~\ref{tab:vqa_results} shows that JOVE-ICE reduces latency and token use relative to FR. TTFT falls by approximately \textbf{13\%} on both benchmarks, while E2E latency decreases by \textbf{7.2\%} on SuperGlasses and \textbf{10.5\%} on WearVQA. Token use drops by \textbf{14.8\%} and \textbf{11.6\%}, respectively. These gains come with modest accuracy declines of 2.97\% and 1.36\%, mainly because implicitly visual questions may be misclassified as non-visual. For example, “Why do cacti have different shapes?” may be routed without requesting image input, causing the response to lack visual evidence and diverge from the ground truth. Despite this trade-off, JOVE-ICE’s accuracy remains comparable to the baseline.

\begin{table}[t]
\centering
\small
\setlength{\tabcolsep}{4pt}
\caption{Visual question answering results on SuperGlasses and WearVQA. Boldface marks the best value in each column.}
\label{tab:vqa_results}
\resizebox{\linewidth}{!}{%
\begin{tabular}{@{}l*{8}{c}@{}}
\toprule
& \multicolumn{4}{c}{SuperGlasses}
& \multicolumn{4}{c}{WearVQA} \\
\cmidrule(lr){2-5}\cmidrule(lr){6-9}
Method
& Acc. (\%) & TTFT (s) & E2E (s) & Total tokens
& Acc. (\%) & TTFT (s) & E2E (s) & Total tokens \\
\midrule
FR
& \textbf{45.46} & 4.5251 & 6.6057 & 4,575.8
& \textbf{51.88} & 2.2942 & 3.2070 & 2,641.0 \\
\rowcolor[RGB]{235,244,252}
JOVE-ICE
& 42.49 & \textbf{3.9065} & \textbf{6.1277} & \textbf{3,897.4}
& 50.52 & \textbf{1.9946} & \textbf{2.8711} & \textbf{2,334.8} \\
\bottomrule
\end{tabular}%
}
\end{table}

\begin{table}[t]
\centering
\small
\setlength{\tabcolsep}{4pt}
\caption{Text-only interaction results on WearVox. The Mean column reports the average across the five tasks. Boldface marks the best value for each metric in the Mean column.}
\label{tab:wearvox_text_results}
\resizebox{\linewidth}{!}{%
\begin{tabular}{@{}ll*{6}{c}@{}}
\toprule
Method & Metric & \shortstack{Search-\\grounded QA} & \shortstack{Closed-\\book QA} & \shortstack{Side-talk\\rejection} & \shortstack{Tool\\calling} & \shortstack{Dialogue\\translation} & Mean \\
\midrule
FR & Task score & 47.90 & 41.50 & 70.90 & 2.49 & 82.01 & 48.96 \\
 & TTFT (s) & 12.7471 & 4.6094 & 6.7728 & 7.5702 & 2.0594 & 6.7518 \\
 & E2E (s) & 15.5791 & 5.0010 & 8.3289 & 9.9487 & 13.9773 & 10.5670 \\
 & Total tokens & 21710.5 & 2859.3 & 9859.4 & 10819.5 & 2320.2 & 9513.8 \\
\addlinespace
\rowcolor[RGB]{235,244,252}
JOVE-ICE & Task score & 46.44 & 40.65 & 72.30 & 20.18 & 76.66 & \textbf{51.25} \\
\rowcolor[RGB]{235,244,252}
 & TTFT (s) & 13.1305 & 4.4700 & 3.9231 & 5.3121 & 1.7782 & \textbf{5.7228} \\
\rowcolor[RGB]{235,244,252}
 & E2E (s) & 15.3773 & 4.7604 & 5.3264 & 7.3164 & 12.3877 & \textbf{9.0336} \\
\rowcolor[RGB]{235,244,252}
 & Total tokens & 21720.1 & 2356.7 & 7565.4 & 8617.8 & 1945.2 & \textbf{8441.0} \\
\bottomrule
\end{tabular}%
}
\end{table}

\begin{table}[t]
\centering
\small
\setlength{\tabcolsep}{4pt}
\caption{Mean classifier GPU runtimes and estimated parallel
classification latency across the three datasets. All values are
in \textbf{milliseconds}.}
\label{tab:classifier_latency}
\begin{tabular}{@{}lrrr@{}}
\toprule
Classifier & SuperGlasses & WearVQA & WearVox \\
\midrule
Device-control intent & 10.581 & 10.566 & 11.702  \\
Visual requirement    &  6.825 &  6.757 &  7.366  \\
Task complexity       & 12.344 & 12.414 & 13.001  \\
\midrule
Parallel classification & \textbf{12.375} & \textbf{12.459}
                        & \textbf{13.228}  \\
\bottomrule
\end{tabular}
\end{table}  

Table~\ref{tab:wearvox_text_results} shows that JOVE-ICE improves the overall WearVox score from 48.96 to \textbf{51.25}, while reducing TTFT, E2E latency, and token use by \textbf{15.2\%}, \textbf{14.5\%}, and \textbf{11.3\%}, respectively. The largest accuracy gain occurs in tool calling. Because WearVox includes tool descriptions in the input query, JOVE-ICE can route these requests to the Text-based Conversational Subagent, which selects the appropriate tools through text-based reasoning. In contrast, FR routes all requests through the Tool-Augmented Subagent, potentially triggering unnecessary tool calls or complex searches. Avoiding this redundant processing improves both efficiency and accuracy.

Table~\ref{tab:classifier_latency} reports the latency of the neural intent classifiers. Across all three datasets, parallel classification takes approximately 12–13 ms per request, adding only minimal overhead to task routing.

\subsection{Session Management Evaluation}
This experiment evaluates our intent-guided session management method on the Session Management Dataset. We submit each query to the module described in \ref{sec:session_mgt}, which determines whether to start a new session or continue the current one. For the soft trigger in Eq.~\ref{equ:soft_trigger}, we set \(w_t=1\), \(w_d=60\), and \(\tau=120\) seconds. We treat stop keywords and device-control queries as hard triggers.

Given the class imbalance (82.9\% Continue vs. 17.1\% Split), accuracy alone may obscure boundary-detection performance. We therefore report per-class precision, recall, and F1-score (Table~\ref{tab:session_metrics}). The confusion matrix is shown in Table~\ref{tab:confusion}.

  \begin{table}[t]
  \centering
  \small
  \begin{minipage}[t]{0.48\linewidth}
  \centering
  \caption{Confusion matrix ($N = 199$).}
  \label{tab:confusion}
  \begin{tabular}{lcc}
  \hline
  & \textbf{Pred. Split} & \textbf{Pred. Cont.} \\ \hline
  \textbf{Actual Split}    & 33  & 1   \\
  \textbf{Actual Continue} & 2   & 163 \\ \hline
  \end{tabular}
  \end{minipage}
  \hfill
  \begin{minipage}[t]{0.48\linewidth}
  \centering
  \caption{Per-class metrics.}
  \label{tab:session_metrics}
  \begin{tabular}{lcc}
  \hline
  \textbf{Metric} & \textbf{Split} & \textbf{Continue} \\ \hline
  Precision & 94.29\% & 99.39\% \\
  Recall    & 97.06\% & 98.79\% \\
  F1-score  & 95.65\% & 99.09\% \\ \hline
  \end{tabular}
  \end{minipage}
  \end{table}

For the \textit{Continue} class, the method achieves 99.39\% precision and 98.79\% recall (163/165), correctly identifying nearly all turns that continue an existing session. For \textit{Split}, it achieves 97.06\% recall (33/34), missing only one boundary: “open translator,” which was classified as a continuation. There are also two false positives, including the context-dependent follow-up, such as “China Telecom, is there any hotel nearby?”, which may have been mistaken for a standalone request because of the abrupt topic shift. Overall, the results show that the method rarely misses genuine boundaries and its false splits are limited to ambiguous queries.

\section{Conclusion}
This paper presents JOVE-ICE, a cloud–edge–device collaborative framework for AI glasses. Across public benchmarks, JOVE-ICE reduces time to first token, end-to-end latency, and token usage compared with the baseline. Analysis of real-world requests shows that many routine device-control tasks can be handled locally, reducing reliance on cloud services and their associated costs. Together, these findings highlight the value of matching each request to an appropriate execution resource. JOVE-ICE offers a practical foundation for scaling wearable assistants across diverse tasks, with a flexible architecture that can accommodate new capabilities as user needs evolve.

\bibliographystyle{splncs04}
\bibliography{mybibliography}
\end{document}